%% file: acl_FINANCEDOMAIN.tex
\documentclass[10pt, a4paper]{article}

\usepackage[final]{lrec2026}

\usepackage{times}
\usepackage{latexsym}
\usepackage{comment}
\usepackage[T1]{fontenc}

\usepackage[english]{babel}
\usepackage[utf8]{inputenc}

\usepackage{microtype}
\usepackage{enumitem}
\setlist{nolistsep}

\usepackage{inconsolata}

\usepackage{alltt}

\usepackage{graphicx}
\usepackage{soul}

\usepackage{booktabs, multirow} 
\usepackage{soul}
\usepackage{xcolor,colortbl} 
\usepackage{changepage,threeparttable} 
\usepackage{longtable}
\usepackage[normalem]{ulem}
\usepackage{lscape}
\usepackage{pifont}
\usepackage{xspace}
\useunder{\uline}{\ul}{}
\usepackage{adjustbox}

\newcommand{\ScribeFinance}{{\sc Scribe Finance}\xspace}

\newcommand\anna[1]{{\color{magenta}#1}}
\newcommand\newtext[1]{{\color{blue!70}#1}}
\newcommand\todo[1]{{\color{red}#1}}

\input{draft_final}

\title{When Tables Go Crazy: Evaluating Multimodal Models on French Financial Documents}

\name{Virginie Mouilleron$^1$ \quad Théo Lasnier$^{1,2}$ \quad  Anna Mosolova$^1$ \quad Djamé Seddah$^1$} 

\address{$^1$ Inria Paris, France \\
         $^2$ Sorbonne Université, Paris, France\\  
         \href{mailto:virginie.a.mouilleron@inria.fr}{\{virginie.a.mouilleron, theo.lasnier, anna.mosolova, djame.seddah\} @inria.fr}\\}

\abstract{
Vision-language models (VLMs) perform well on many document understanding tasks, yet their reliability in specialized, non-English domains remains underexplored. This gap is especially critical in finance, where documents mix dense regulatory text, numerical tables, and visual charts, and where extraction errors can have real-world consequences.
We introduce \ScribeFinance, the first multimodal benchmark for evaluating French financial document understanding. The dataset contains 1,204 expert-validated questions spanning text extraction, table comprehension, chart interpretation, and multi-turn conversational reasoning, drawn from real investment prospectuses, KIDs, and PRIIPs.
We evaluate six open-weight VLMs (8B--124B parameters) using an LLM-as-judge protocol. While models achieve strong performance on text and table tasks (85--90\% accuracy), they struggle with chart interpretation (34--62\%). Most notably, multi-turn dialogue reveals a sharp failure mode: early mistakes propagate across turns, driving accuracy down to roughly 50\% regardless of model size. \\
These results show that current VLMs are effective for well-defined extraction tasks but remain brittle in interactive, multi-step financial analysis. \ScribeFinance offers a challenging benchmark to measure and drive progress in this high-stakes setting.
\\ \newline 
}

\begin{document}

\maketitleabstract


\section{Introduction}
\begin{figure*}[!ht]
    \centering
    \includegraphics[width=1\textwidth]{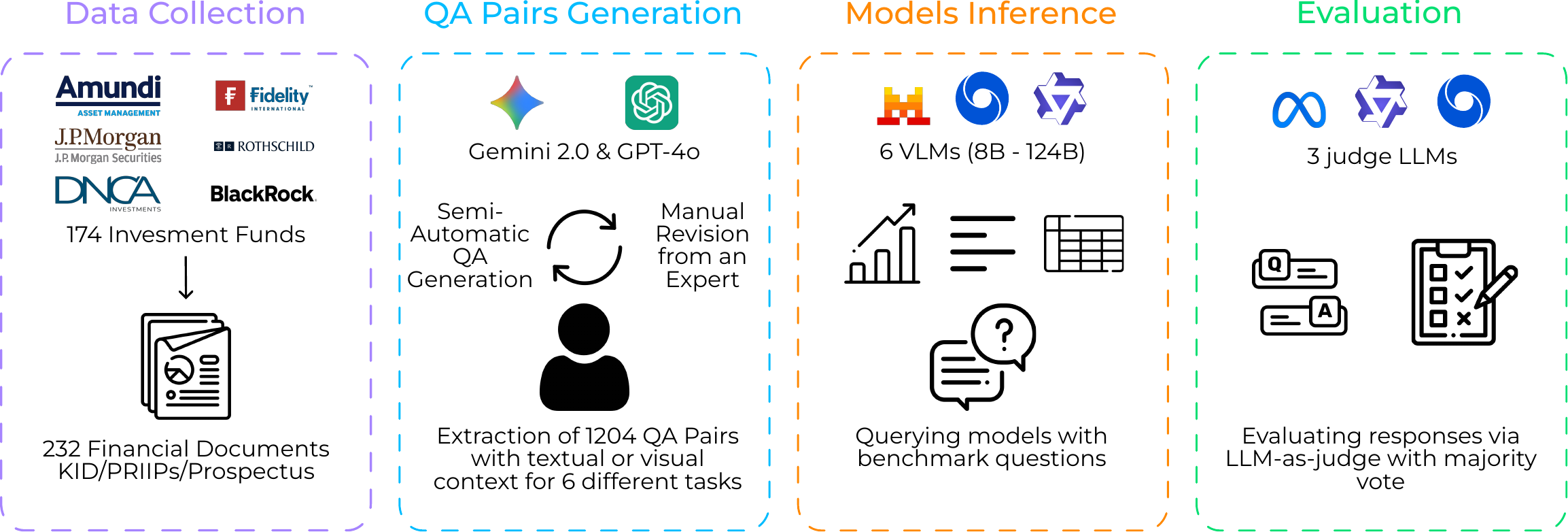}
    \caption{Overview of the \ScribeFinance benchmark construction and evaluation pipeline. French financial documents (prospectuses, KIDs, PRIIPs) are collected from  asset management companies, then processed to generate question-answer pairs spanning text, tables, and charts. Six Vision-Language Models are evaluated on these tasks, with responses assessed using a majority-vote LLM-as-judge protocol.}
    \label{fig:pipeline}
\end{figure*}

The 2008--2009 global financial crisis exposed major failures in transparency and regulatory oversight across financial markets, prompting a coordinated international response to strengthen disclosure and investor protection requirements \cite{g20_2009_london_statement}. In the European Union, these reforms were subsequently formalized through regulations such as \textit{Markets in Financial Instruments Directive (MiFID II)} and \textit{Packaged Retail and Insurance-based Investment Products (PRIIPs)}, requiring asset management companies to issue standardized prospectuses at the beginning of the fiscal year and subjecting them to end-of-year review by regulatory authorities \cite{ec_2014_mifid2,eu_2014_priips}.
Following the implementation of these regulations, the volume of regulated disclosure documents has become substantial in all  major financial jurisdictions. In the European Union and the United States alone, tens of thousands of prospectuses and prospectus-like documents, including amendments and standardized investor disclosures, are produced each year, illustrating the scale of financial documentation that must be reviewed and interpreted.

Because of their unprecedented level of performance in many text understanding tasks \citep{grattafiori2024llama3herdmodels}, Large Language Models (LLMs) have become the central component of the  modern Natural Language Processing (NLP) arsenal.  Despite this progress, their evaluation in certain specialized domains remains uneven. Finance, for example, presents several particular challenges: documents are long, terminology is technical, and information is often distributed across text, tables, and charts. Moreover,  most evaluation resources focus on English, leaving models' capabilities in other languages, particularly in domain-specific contexts, largely untested. This gap is especially problematic for regulatory and advisory applications where extraction accuracy is critical: a financial advisor querying a prospectus for the entry fee of a specific share class cannot tolerate hallucinated percentages.

French financial documents exemplify these challenges. Investment prospectuses, which describe potential returns and risks of financial products, can span 10 to 600+ pages and combine dense legal prose with complex tabular data and visual elements. Deploying LLMs in such challenging scenarios requires rigorous evaluation of their ability to locate and extract precise information, a prerequisite for higher-level tasks like summarization or compliance verification. 

\draftreplace{TODO remove?: However, no dedicated French-language benchmark exists for this purpose}{Despite several efforts to build specialized benchmarks in French for the financial domain \citep{faysse2025colpaliefficientdocumentretrieval, xue2025fammabenchmarkfinancialdomain}, the proposed datasets remain small in scale and limited in coverage ($\approx$ 200 examples; see Table \ref{tab:datasets_comparison})}, making it hard to assess whether state-of-the-art models are ready for these high-stakes applications.

To address this gap, we introduce \ScribeFinance, a multimodal
benchmark dataset of 1,204 questions designed to evaluate VLMs on
French financial document understanding. The dataset spans multiple
question types (open-ended, Yes/No, True/False, and multi-turn
conversational) and input modalities (text, tables, and
charts). Questions range from named entity extraction to complex
reasoning requiring integration of information across document
sections. Figure~\ref{fig:pipeline} provides an overview of the 
benchmark construction and evaluation pipeline.

We evaluate six open-weight, state-of-the-art Vision-Language Models (VLMs) from three model families, spanning scales from 8B to 124B parameters, using an LLM-as-judge evaluation protocol. Results show that while models perform well on text-based questions ($\sim$88--90\%) and achieve moderate to strong performance on table comprehension ($\sim$52--86\%), chart interpretation remains challenging across all models ($\sim$34--62\%).
More critically, the multi-turn conversational task reveals a systematic failure mode: errors propagate across dialogue turns, causing accuracy to collapse to approximately 50\% ($\sim$46--59\%) regardless of model size. This behavior raises concerns about the reliability of current VLMs in interactive financial analysis settings.

Together, these results suggest that while VLMs are effective for well-scoped information extraction, they remain fragile when reasoning must be maintained across visual modalities and conversational context. \ScribeFinance provides a benchmark for quantifying these limitations and tracking progress on French financial document understanding.
Our main contributions are:
\begin{itemize}
\item \ScribeFinance, the first multimodal benchmark for French financial document understanding, consists of 1,204 expert-validated questions spanning text extraction, table comprehension, chart interpretation, and multi-turn dialogue.\footnote{The dataset and its accompanying resources can be accessed here \url{https://github.com/dseddah/Scribe_finance/}}
\item A systematic evaluation of six state-of-the-art VLMs, showing strong performance on text and tables but persistent weaknesses in chart interpretation and conversational settings.
\item Empirical evidence that error propagation in multi-turn dialogue negates scaling benefits, with model accuracy converging to approximately 50\% regardless of parameter count.
\end{itemize}

\section{Related Works}

\subsection{French Evaluation Resources}

Most NLP evaluation benchmarks target English, but some efforts have introduced resources for French-language evaluation as well. General-purpose question answering benchmarks, such as FQuAD \cite{dhoffschmidt-etal-2020-fquad} and PIAF \cite{keraron_project_2020}, largely derived from Wikipedia and inspired by their English counterpart SQuAD \cite{rajpurkar2016squad}, have played an important role in enabling French-language QA evaluation. More recent efforts have extended evaluation beyond general domains, including FrenchMedMCQA \cite{labrak_frenchmedmcqa_2023} for medical reasoning and French CrowS-Pairs \cite{neveol_french_2022,nangia_crows-pairs_2020} for bias assessment. Additional datasets such as Newsquadfr\footnote{\url{https://huggingface.co/datasets/lincoln/newsquadfr}} further explore model performance on journalistic and informal French content.

This section does not attempt to provide an exhaustive survey of French evaluation datasets. Instead, these resources illustrate that, despite growing coverage of French language understanding, existing benchmarks largely focus on short, text-only inputs and general or domain-specific knowledge. In contrast, our work targets multimodal, long-form financial documents and evaluates model behavior in high-stakes, document-centric settings.

\subsection{NLP Work in the Finance Domain}

\paragraph{General-purpose Multilingual Benchmarks Including French}
While English-centric evaluation remains the norm, several multilingual benchmarks provide partial coverage of French. Datasets such as MKQA \cite{longpre_mkqa_2021}, XQA \cite{liu_xqa_2019}, and MIRACL \citep{zhang_miracl_2023} provide cross-lingual question-answering benchmarks primarily based on Wikipedia, enabling evaluation of multilingual transfer across a range of languages, including French. These resources have played an important role in advancing multilingual evaluation, but they focus on short, text-only inputs and do not address the challenges posed by long, structured, or domain-specific documents.


\paragraph{Specialized Financial Benchmarks in English}
The financial domain has also motivated the development of specialized benchmarks targeting numerical and document-level reasoning. TAT-QA \citep{zhu_tat-qa_2021} and FinQA \citep{chen_finqa_2022} evaluate reasoning over financial reports by combining textual passages with tabular data, requiring models to perform arithmetic and logical operations rather than simple extraction. More recent datasets such as ConvFinQA \cite{chen_convfinqa_2022} and PACIFIC \cite{deng_pacific_2023} extend this setting to multi-turn conversational scenarios, exposing the challenges of numerical reasoning and context tracking in dialogue-based interactions. Very recently, \citet{lithgow-serrano-etal-2025-assessing} introduced a banking-domain retrieval-augmented generation benchmark focusing on full documents comprising approximately 600 question-answer pairs. 

\paragraph{Specialized Financial Benchmarks in French}

Recently, \citet{faysse2025colpaliefficientdocumentretrieval} shared a small dataset focusing on answering questions partly about financial tables in French documents. In parallel, FAMMA \cite{xue2025fammabenchmarkfinancialdomain} presents a multilingual containing 9\% of french content, multimodal financial benchmark derived from university-level instructional and assessment materials across eight core finance areas, requiring joint reasoning over text, tables, and charts, and proving challenging even for strong models. See Table~\ref{tab:datasets_comparison} for a detailed comparison of these datasets.

\begin{table}[ht]
    \centering
    \begin{adjustbox}{max width=\columnwidth}
    \begin{tabular}{cccccc}
    \hline
    \textbf{Dataset} & \textbf{Size} & \textbf{Question type} & \textbf{Context type} \\
    \hline
       \citet{faysse2025colpaliefficientdocumentretrieval} & 210 & Retrieval & Table  \\
       \citet{xue2025fammabenchmarkfinancialdomain} & 190 & Open, MCQ & Table, None  \\
     \multirow{2}{*}{Ours} & \multirow{2}{*}{1,204} & Open, MCQ,  & \multirow{2}{*}{Table, Chart, Text}  \\
      &  &  TFQ, Yes/No &   \\
     \hline
    \end{tabular}
    \end{adjustbox}
    \caption{Comparison between existing financial benchmarks and our newly proposed \ScribeFinance (see Section \ref{sec:newdataset}). \textit{MCQ} = multiple-choice questions, \textit{TFQ} = True/False questions, \textit{Yes/No} = Yes/No questions.}
    \label{tab:datasets_comparison}
\end{table}

Despite these advances, existing financial benchmarks remain limited in several respects: they are predominantly English-only, focus on relatively short excerpts rather than full-length regulatory documents, and largely exclude multimodal inputs such as charts. In contrast, our work targets French financial prospectuses, which are long, multimodal, and legally constrained, and evaluates model behavior in high-stakes document understanding and conversational settings.

\input{table_tab_distribution}

\section{Designing \ScribeFinance}
\label{sec:newdataset}

 

Building \ScribeFinance required balancing realism, scale, and annotation reliability. French financial prospectuses are long, highly structured, and repetitive  documents that combine dense legal text with tables and charts, frequently spanning hundreds of pages. Rather than treating these documents as monolithic inputs, we extract excerpts of varying lengths (0.5-30 pages) and focus on evaluating a model’s ability to accurately locate and extract specific, document-grounded information, which is a prerequisite for reliable downstream reasoning in financial settings.

The dataset was constructed from publicly available French financial documents collected from multiple asset management companies, including \textit{prospectuses, Key Information Documents (KIDs)}, and \textit{Packaged Retail and Insurance-based Investment Products (PRIIPs)} published over the past 15 years.\footnote{The documents were collected from asset management companies and are publicly available under EU and U.S. financial regulations \cite{ec_2014_mifid2,eu_2014_priips}, which require publication for investor protection and public transparency. They are accessible without authentication or paywalls on official issuer or regulator websites and contain no private or personally identifiable information.}

In the next section, the approach to generate questions for text-based and image-based tasks is described in detail.

\subsection{Question Generation}

Question construction followed an iterative, semi-automatic process designed to identify salient, extractable financial information. Two LLMs (GPT-4o and Gemini-2.0) assisted in generating candidate question and answers, after which all outputs were reviewed and revised by a human annotator. When necessary, input contexts were expanded to ensure completeness and faithful grounding in the source documents. The specific design choices for each question type are described below.

\paragraph{Text-Based Task} 

Text-based questions were derived from PDF documents converted semi-automatically into text. During this process, tables were preserved and transformed into tabulated textual format, resulting in contexts combining plain text and structured tables. This subset primarily focuses on extracting key financial information, such as applicable taxes and minimum investment durations. 


To assess the suitability of LLMs for question generation in this task, we first conducted a preliminary analysis to determine whether salient and informative content could be reliably extracted from the source documents. As the result were satisfactory, we proceeded with the creation of question-answers pairs via prompting. The model was asked to identify twenty key informational items that could form the basis for potential questions, link each item to its textual extract, and generate an open-ended question grounded in that excerpt. All outputs were subsequently validated and, when necessary, rewritten by a human annotator.

\paragraph{Image-Based Tasks: Tables and Charts}

For table- and chart-centered tasks, questions and answers were generated directly from visual inputs. This subset includes open-ended, Yes/No, and True/False questions. The objective is to evaluate structured and graphical data interpretation in financial documents.

\paragraph{Image-Based Task: Conversational Setting} 

The conversation-based subset (referred to as \textit{Conv.} in the tables) targets multi-step reasoning over financial content. Unlike the rest of the dataset, where answers are directly extractable, these questions involve mathematical reasoning and are formatted as multiple-choice questions. This subset was generated using a dedicated prompt specifying both the structure of the conversational turns and the syntactic diversity, as well as the nature of the references to be included in the dialogue. This design enables controlled evaluation of error propagation in interactive settings.

\begin{table*}[ht!]
    \centering
    {\footnotesize
    \begin{tabular}{p{2.99cm}p{12.5cm}}
\hline
\textbf{Question Type} & \multicolumn{1}{c}{\textbf{Example}} \\
\hline
Text Question & \emph{«~Ce fonds est-il géré activement ou suit-il un indice de manière passive ?~»} \\
Table Comprehension & \emph{«~À combien s'élèvent les frais courants annuels prélevés par le FCPE ?~»} \\
Chart Interpretation &  \emph{«~Combien de périodes consécutives sans cristallisation sont visibles sur le graphique ?~»}  \\
Special Cases & \emph{«~Quels instruments dérivés spécifiques peuvent être utilisés par le compartiment ?~»} \\
Conversational & 1\textsuperscript{st} turn: \emph{«~Si je place 25 000 € sur la part A, combien me coûteraient les frais d’entrée maximum ?~»} \\ 
 & 2\textsuperscript{nd} turn:
\emph{«~Et si je prends cette même somme pour la part I ou R, j’aurais une différence au niveau des frais ?~»} \\
 \hline
    \end{tabular}
    }
    \caption{Examples of each question type in the \ScribeFinance dataset. Visual examples are provided in Appendix \ref{app:dataset_examples} and English translations in Appendix \ref{tab:transation_scribe_examples}.}
    \label{tab:scribe_examples}
\end{table*}

\subsection{Manual Validation and Refinement}

All question–answer pairs were validated by a French financial domain expert.\footnote{The expert annotator has two years of experience as a financial data scientist leading a financial data annotation team. As each generated question had a single directly verifiable answer, one expert was deemed sufficient for dataset verification and rewriting.} Each question was assigned a gold-standard answer confirmed by the expert. 

The validation process covered question formulation, answer correctness, and manual verification of all document excerpts used as inputs, with particular attention to open-ended questions. Instances were removed if answers were incorrect, questions were overly generic, insufficiently grounded in the source table, too short, or repetitive. To increase linguistic and structural diversity, a substantial subset of the remaining questions was rewritten by the same annotator. In total, 75\% of questions were reformulated or removed by the annotator.

\subsection{Task Overview and Dataset Composition}

The benchmark comprises six task categories reflecting realistic financial document understanding scenarios. Except for text-only questions, all tasks involve multimodal inputs, where a relevant image (e.g., a table, chart, or document page) is provided alongside the textual context.

The task categories (examples in Table \ref{tab:scribe_examples}) are: 
\begin{itemize}
    \item \textbf{Text Question}, focusing on extraction from purely textual contexts;
    \item \textbf{Table Comprehension}, requiring reasoning over structured tabular data; 
    \item \textbf{Chart Interpretation}, based on graphical financial representations;
    \item \textbf{Special Cases}, involving nuanced terminology or implicit reasoning;
    \item \textbf{Conversational (Gold Context)}, with a dialogue with oracle previous answers;
    \item \textbf{Conversational (Model Context)}, with a dialogue with model-generated previous answers to study error propagation.
\end{itemize}  


To capture a range of retrieval and reasoning challenges, tasks vary along two 
axes: \textbf{context length}, ranging from short excerpts to document-level inputs, and \textbf{context modality}, including plain text, tables, charts, and mixed formats. Questions are formulated as open-ended, binary (Yes/No, True/False), or multiple-choice depending on the task.

\subsection{Dataset Statistics}
Table~\ref{tab:distribution} summarizes the distribution of questions across task categories, question types, and context modalities. The dataset includes both text-based and image-based questions, with open-ended formats dominating overall, while binary and conversational formats target more constrained reasoning settings.

Context lengths range from short passages (1--2 sentences) to multi-page documents. The conversational subset consists of 5--10 turn dialogues, explicitly designed to probe error propagation and robustness in interactive scenarios.

\section{Experimental Setup}

\paragraph{Models} We evaluated six state-of-the-art Vision-Language Models spanning different scales and architectures: Qwen/Qwen3-VL-8B-Instruct and Qwen/Qwen3-VL-32B-Instruct~\cite{qwen3technicalreport}, google/gemma-3-12b-it and google/gemma-3-27b-it~\cite{gemma_2025}, and mistralai/Pixtral-12B-2409 and mistralai/Pixtral-Large-Instruct-2411~\cite{agrawal2024pixtral12b}. Model sizes range from 8B to 124B parameters, enabling analysis of scaling effects on financial document understanding.

\paragraph{Answer Generation} For each task, models received a prompt and were instructed to respond concisely without explanations. Single turn image-based tasks (Table, Charts, Special Cases) included the image followed by the question. The Text Question task provided textual context instead of an image. Conversational tasks built a multi-turn dialogue incrementally, with the image provided only at the first turn and subsequent model responses appended to the history. In all cases, the assistant turn was prefilled with ``Answer:'' to constrain the response format. We used greedy decoding to ensure reproducibility. Complete prompt templates are provided in Appendix~\ref{app:prompt_template}.

\paragraph{Evaluation Protocol} Given that more than a half of the proposed dataset consists of open-ended questions (Table \ref{tab:distribution}), and to ensure a unified evaluation process across all tasks, we adopt an LLM-as-judge approach\footnote{We initially attempted to extract answers automatically using regular expressions for certain question types, however this approach proved error-prone, so we switched to the LLM-as-judge method.} to avoid the high cost of human validation \citep{Zheng2023llmasajudge}. We used three open-source judge models\footnote{Only open-source models were used, in compliance with the restrictions established by our institution.} independently to assess each response: meta-llama/Llama-3.3-70B-Instruct ~\cite{grattafiori2024llama3herdmodels}, Qwen/Qwen3-32B~\cite{qwen3technicalreport}, and google/gemma-3-27b-it~\cite{gemma_2025}\footnote{Although Qwen and Gemma family models are used both for answer generation and evaluation, which may raise concerns about self-preference bias, \citet{chen-etal-2025-beyond} show that models larger than 7B parameters exhibit limited self-bias and that the strongest self-preference effects are observed in the Llama family, which in our setup is used only for the evaluation.}. An answer was considered correct if a majority of judges determined it to be correct. Scores reported in Table~\ref{tab:agg_results} represent the percentage of questions answered correctly under this majority-vote criterion.

\section{Results and Analysis}
\begin{table*}[!ht]
\centering
{\footnotesize
\begin{tabular}{l|rrrrrr|r}
\hline

\multirow{2}{*}{\textbf{Model}}  & \multicolumn{6}{c|}{\textbf{Task}} & \multicolumn{1}{l}{\multirow{2}{*}{\textbf{Avg.}}} \\
   & \multicolumn{1}{l}{\textbf{Text}} &
  \multicolumn{1}{l}{\textbf{Tables}} &
  \multicolumn{1}{l}{\textbf{Charts}} &
  \multicolumn{1}{l}{\textbf{Conv. Gold}} &
  \multicolumn{1}{l}{\textbf{Conv.}} &
  \multicolumn{1}{l|}{\textbf{Special Cases}} &
   \\ \hline
Qwen3-VL-8B  & {\ul 89.4}    & 80.0          & 45.3          & {\ul 73.8}    & {\ul 52.3}    & {\ul 63.6}    & {\ul 67.8}    \\
Gemma-3-12B  & 88.0          & \textbf{85.8} & 46.1          & 70.8          & 50.8          & 40.9          & 63.8         \\
Pixtral-12B  & 88.4          & 51.7          & 34.4          & 63.1          & {\ul 52.3}    & 27.3          & 53.4          \\
Gemma-3-27B  & 89.0          & {\ul 85.0}    & {\ul 48.4}    & {\ul 73.8 }   & 49.2          & 54.5          & 66.2          \\
Qwen3-VL-32B & \textbf{89.8} & \textbf{85.8} & \textbf{61.7} & \textbf{86.2} & \textbf{58.5} & \textbf{72.7} & \textbf{75.6} \\ 
Pixtral-Large-124B & 87.8         & 71.7          & 46.1          & 70.8          & 46.2          & {\ul 63.6}  & 55.2          \\ 
\hline
\end{tabular}
}
\caption{Model performance on \ScribeFinance (accuracy \%). Text and table tasks achieve strong results (80--90\%), while chart interpretation (34--62\%) and multi-turn conversation (46--59\%) display significant weaknesses. \textbf{Bold} indicates best performance; \underline{underline} indicates second best.}

\label{tab:agg_results}
\end{table*}

\begin{figure*}[!ht]
    \centering
    \includegraphics[width=0.9\textwidth]{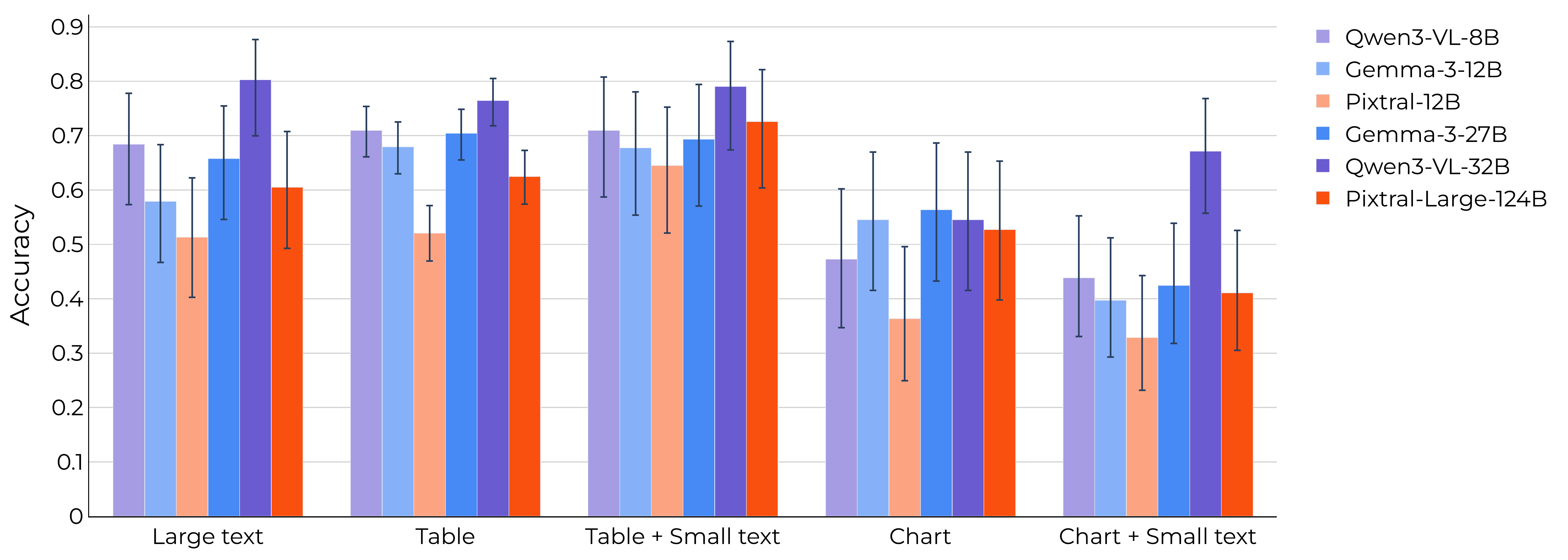}
    \caption{Model accuracy on image-based question subcategories. Performance remains strong on table comprehension tasks (70--86\%) but degrades substantially on chart interpretation (34--62\%). Qwen3-VL-32B consistently outperforms other models across all visual modalities.}
\label{fig:image_accuracy}
\end{figure*}

\begin{figure*}[!ht]
    \centering
    \includegraphics[width=0.9\textwidth]{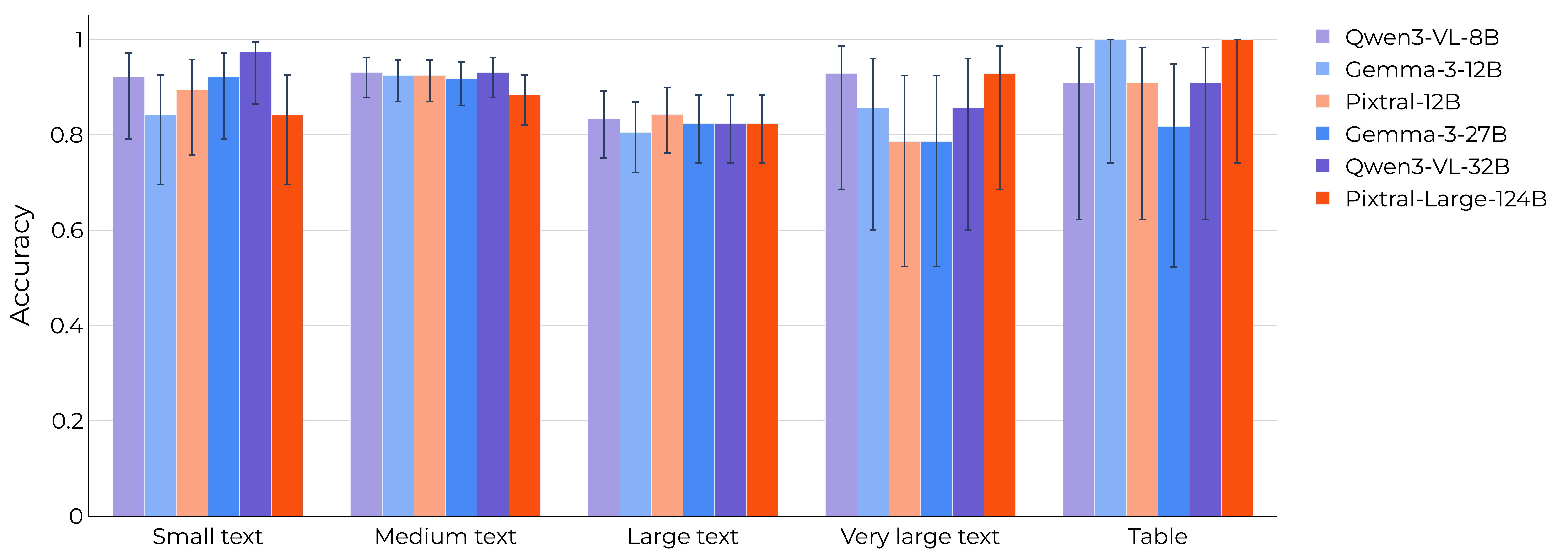}
    \caption{Model accuracy on text-based question subcategories by context length. All models achieve high performance (85--95\%) on short and medium text contexts, with moderate degradation on larger contexts. Performance on tabular text (rightmost) remains competitive, indicating that text-based table comprehension is less challenging than image-based table interpretation.}
\label{fig:text_accuracy}
\end{figure*}

Table~\ref{tab:agg_results} presents results across all task
categories. Qwen3-VL-32B achieved the strongest overall performance
with an average score of 75.6\%, obtaining top scores across all six
categories. Qwen3-VL-8B followed with 67.8\%, Gemma-3-27B reached
66.2\%, Gemma-3-12B scored 63.8\%, Pixtral-Large-124B achieved 55.2\%,
and Pixtral-12B showed the lowest performance at 53.4\%.

For most tasks, models achieved strong performance in the 70-90\%
range. Text Question scores approached 90\% across all models, and
table comprehension reached 85.8\% for the best performers, indicating
that current Vision-Language Models handle both textual entity
extraction and structured visual information effectively when the task
is well-defined.

Figure~\ref{fig:text_accuracy}  shows that text-based accuracy remains high across short and medium contexts, with only moderate degradation as context length increases.

Two tasks exposed notable limitations. First, chart interpretation is
challenging for all tested models, with scores ranging from 34.4\%
(Pixtral-12B) to 61.7\% (Qwen3-VL-32B). The second best-performing
model does not reach 50\%. Though charts are designed to render complex
information more accessible to human understanding, this visual
simplification paradoxically challenges our tested models, which struggle
to extract trends, comparisons, and proportions from graphical
elements rather than explicit text or tabular image.

Figure~\ref{fig:image_accuracy} provides a fine-grained breakdown of image-based performance, showing strong results on table comprehension but a substantial drop on chart-based questions across all models.

Second, the conversational task evaluation showed how error propagation affects
multi-turn reasoning. In the gold context condition, where correct
previous answers are provided, models achieved 63.1--86.2\% accuracy
with clear differentiation by model capacity. In the standard
condition, where models must build on their own previous responses,
performance dropped sharply and converged to a narrow 46.2--58.5\%
range. The comparison between the Conversational Gold and
Conversational Standard task suggests that the bottleneck is not
reasoning capacity per se, but rather the accumulation of errors
across turns: once a model makes an early mistake, subsequent answers
are compromised by incorrect context, and larger models offer no
protection against this cascade. These findings question
the reliability of VLMs in interactive, multi-turn
financial analysis scenarios where accumulated errors cannot be
corrected.




\input{new_Discussion}

\section{Conclusion}
\label{sec:conclusion}
We introduced \ScribeFinance, a multimodal benchmark for
evaluating Vision-Language Models on French financial document
understanding. The benchmark targets realistic, high-stakes scenarios
involving long, heterogeneous documents that combine legal text,
numerical tables, and charts, and includes both single-turn and
multi-turn conversational tasks. By focusing on excerpt-grounded
information extraction rather than full-document access,
\ScribeFinance emphasizes precise retrieval as a prerequisite for
reliable financial reasoning.

Our evaluation of six state-of-the-art VLMs presents a clear
contrast between strong performance on well-scoped text and table
extraction tasks and persistent weaknesses in chart interpretation and
conversational settings. In particular, we show that error propagation
across dialogue turns causes model performance to collapse regardless
of scale, exposing a failure mode that is largely invisible in
standard single-turn benchmarks.

Together, these findings suggest that progress in financial document
understanding will require advances beyond model scaling alone. Future
work should explore training objectives and architectural mechanisms
that explicitly support uncertainty awareness, error correction, and
robust multi-step reasoning over multimodal inputs. We hope that
\ScribeFinance will serve as a useful testbed for measuring such
progress and for guiding the development of more reliable models for
real-world financial analysis.

\section{Limitations}
\input{Limitations}

\section{Ethics}
All documents used in this study are publicly available financial disclosures released by asset management companies. No private, sensitive, or personally identifiable information was collected or processed. The expert reviewer involved in dataset construction and evaluation was fairly compensated for their contributions.
While the benchmark is designed for document understanding and evaluation purposes, financial applications are inherently high-stakes. Our results identify failure modes such as error propagation in conversational settings, exposing the risk of over-reliance on automated systems for financial analysis or advisory tasks. The benchmark is intended to support research and evaluation, not to replace professional judgment in real-world financial decision-making.

\section{Acknowledgments}

We thank the reviewers for their valuable feedback on our work. We are grateful for all the comments and feedback from Iacopo Poli, Oskar Hallström, and Adrien Cavaillès from LightOn on an earlier version of the \ScribeFinance dataset. 

This work was performed using HPC resources from GENCI–IDRIS (Grant 2025-AD011016564) on the supercomputer Jean
Zay’s CSL, A100, and H100 partitions. This project was supported by the BPI Code Common and Scribe projects as well as by Djamé Seddah's PRAIRIE-PSAI chair, funded by the French national agency ANR, as part of the “France 2030” strategy under the reference ANR-23-IACL-0008.



\newpage
\section{Bibliographical References}

\bibliographystyle{lrec2026-natbib}
\bibliography{Scribe_Models_evaluation_2}

\newpage

\section{Appendix}

\input{appendix_en}

\end{document}

%% file: draft_final.tex
\expandafter\def\expandafter\UrlBreaks\expandafter{\UrlBreaks
  \do\a\do\b\do\c\do\d\do\e\do\f\do\g\do\h\do\i\do\j%
  \do\k\do\l\do\m\do\n\do\o\do\p\do\q\do\r\do\s\do\t%
  \do\u\do\v\do\w\do\x\do\y\do\z\do\A\do\B\do\C\do\D%
  \do\E\do\F\do\G\do\H\do\I\do\J\do\K\do\L\do\M\do\N%
  \do\O\do\P\do\Q\do\R\do\S\do\T\do\U\do\V\do\W\do\X%
  \do\Y\do\Z}

\usepackage{soulutf8}
\usepackage[normalem]{ulem}
\usepackage{xcolor}

\usepackage{enumitem}
\setlist{nolistsep}
\usepackage{microtype}

\iffalse 
    \usepackage[final]{proofing}
  \renewcommand\hl[1]{{#1}}  
   {\draftnote{\red{#2}}}
   \newcommand\redHL[1]{}
  \renewcommand\todo[1]{}

  \newcommand{\Djame}[1]{}

\else  

\usepackage[final]{proofing}

\newcommand{\Djame}[1]{
\textbf{\textcolor{red}{\hl{Djame: #1}}}
}

\newcommand\red[1]{{{\textcolor{red}{\bf #1}}}}

\let\oldred\red
\renewcommand\red[1]{{ \oldred{{#1}}}}

 \newcommand\redHL[1]{\red{\hl{#1}}}
\let\olddraftnote\draftnote
\renewcommand\draftnote[1]{\olddraftnote{\red{#1}}}

\fi

%% file: table_tab_distribution.tex
\begin{table*}[!ht]
\centering
{\footnotesize
\begin{tabular}{llcccccc}
\hline
 \multicolumn{2}{c}{\multirow{3}{*}{\textbf{Question/Context Type $\downarrow$}}} & \multicolumn{5}{c}{\textbf{Task type $\downarrow$}} \\
\cline{3-7} 
& & \multicolumn{1}{c}{\textbf{Text-Based}} & \multicolumn{4}{c}{\textbf{Image-Based}} \\ \cline{3-7} 
\multicolumn{2}{l}{}                      & \multicolumn{1}{c}{\textbf{Text}}             & \textbf{Tables} & \textbf{Charts} & \textbf{Conv.} & \textbf{Special Case} \\ \hline
\multicolumn{1}{l}{\multirow{4}{*}{Question Type}} & Open                 & 501 & 248 & 94     & 0     & 19            \\
\multicolumn{1}{l}{}                               & Yes/No               & 0   & {213}   & 28     & 0     & 3             \\
\multicolumn{1}{l}{}                               & True/False           & 0   & 27   & 6      & 0     & 0             \\
\multicolumn{1}{l}{}                               & {MCQ}                  & 0   & 0    & 0          & 65    & 0             \\ \hline

\multicolumn{2}{l}{Total (per question type)}                                                 & 501 & {488}  & 128    & 65    & 22            \\ \hline
\\\hline
\multicolumn{1}{l}{\multirow{10}{*}{Context Type}} & Small text           & 38  & 0    & 0          & 0     & 0             \\
\multicolumn{1}{l}{}                               & Medium text          & 146 & 0    & 0          & 0     & 6             \\
\multicolumn{1}{l}{}                               & Large text           & 108 & 0    & 0          & 30    & 16            \\
\multicolumn{1}{l}{}                               & Very large text      & 14  & 0    & 0          & 0     & 0             \\
\multicolumn{1}{l}{}                               & Document-wise (KID)             & 184 & 0    & 0      & 0     & 0             \\
\multicolumn{1}{l}{}                               & Table                & 11  & {442}  & 0      & 15    & 0             \\
\multicolumn{1}{l}{}                               & Table \& Small text  & 0   & {35}   & 0      & 20    & 0             \\
\multicolumn{1}{l}{}                               & Table \& Medium text & 0   & 11   & 0      & 0     & 0             \\
\multicolumn{1}{l}{}                               & Chart                & 0   & 0   & 73     & 0     & 0             \\
\multicolumn{1}{l}{}                               & Chart \& Small text  & 0   & 0    & 55     & 0     & 0             \\ \hline
\multicolumn{2}{l}{Total (per context type)}                                                 & 501 & {488}  & 128    & 65    & 22            \\ \hline
\end{tabular}
}

\caption{Distribution of the \ScribeFinance benchmark (1,204 questions) across question types and context modalities. The dataset spans text-based questions (501 questions) and image-based questions including tables, charts, multi-turn conversations, and special cases, for a total of 703 questions with an associated image. Open-ended questions dominate {(862 questions)}, with binary (Yes/No, True/False) and conversational {MCQ} formats targeting specific reasoning challenges. \textit{Conv.} = multi-turn conversation questions.}
\label{tab:distribution}
\end{table*}

%% file: new_Discussion.tex
\section{Discussion}
\label{sec:discussion}
This work evaluates the capabilities of current Vision-Language Models on French financial document understanding through the \ScribeFinance benchmark. Beyond reporting performance scores, our results reveal several structural limitations that are particularly relevant for high-stakes, real-world deployment.

\subsection{Model Performance and Limitations}

First, the strong performance observed on text-based and table-based tasks suggests that contemporary VLMs are generally reliable when the task is well-scoped and the relevant information is explicitly present in the input. Extraction of named entities, numerical values, and clearly localized facts appears largely solved under these conditions. This aligns with prior findings on document understanding benchmarks \citep{clark2026molmo2} and indicates that scaling and multimodal pretraining have effectively addressed many single-step retrieval problems. 

However, this apparent robustness does not extend to more visually or temporally complex settings. Chart interpretation remains a consistent weakness across all evaluated models, with large performance gaps relative to text and table tasks. Unlike tables, charts require models to infer trends, relative comparisons, and implicit values that are not directly encoded as text. The persistent difficulty observed here suggests that current VLMs rely heavily on surface-level pattern matching rather than deeper visual abstraction, limiting their ability to reason over graphical representations commonly used in financial reporting.

The most striking finding concerns multi-turn conversational evaluation. When models are required to build on their own previous answers, performance collapses to approximately 50\% regardless of model size. This behavior exposes a failure mode that is not visible in single-turn benchmarks: early mistakes introduce incorrect context that subsequent reasoning cannot recover from. Importantly, the comparison with the Conversational Gold setting indicates that this degradation is not primarily due to a lack of reasoning capacity, but rather to error accumulation and context contamination. Scaling the model does not mitigate this effect, suggesting that architectural or training-level changes may be required to support reliable multi-step financial reasoning.

\subsection{Generation Biases and Implications for Dataset Construction}

Our analysis also reveals several systematic tendencies in the semi-automatic question generation process that have implications for both dataset composition and evaluation outcomes. In the open-ended setting, generated questions disproportionately focused on short, easily identifiable facts, such as entry fee percentages or single numerical values. In contrast, more complex information---particularly investment rules or constraints distributed across multiple sections of a document---was less frequently captured. This suggests that current generation pipelines favor information that is locally salient, which may underrepresent questions requiring broader contextual integration.

We further observed limited lexical diversity and originality in a subset of the generated questions. Similar formulations were often reused across documents, resulting in questions that were syntactically correct but insufficiently specific to the source material. A comparable pattern emerged for table-based inputs: even when tables contained structurally rich or nuanced information, generated questions tended to target straightforward value extraction rather than higher-level relationships or constraints. These tendencies required manual revision (described in Section \ref{sec:newdataset}) to ensure adequate coverage of more challenging reasoning scenarios.

In the multiple-choice setting, additional artifacts emerged. Although the model was able to generate candidate distractors, incorrect answer options were frequently implausible, often falling well outside the range of values or concepts presented in the document. Moreover, the correct answer was repeatedly assigned to the same option label, introducing a positional bias that could be exploited during evaluation. Addressing these issues required manual correction of both distractor content and label assignment. Together, these observations underscore current limitations of automated generation methods and reinforce the importance of human oversight when constructing evaluation benchmarks in high-stakes domains such as finance.

These observations have practical implications. Financial analysis often involves iterative questioning, clarification, and dependency on prior answers. The inability of current models to correct or contain earlier errors raises concerns about their suitability for interactive advisory or compliance-related applications, where even small inaccuracies can propagate into significant downstream risks. Our results therefore caution against over-reliance on conversational interfaces for complex financial document analysis without additional safeguards.

Finally, the use of an LLM-as-judge evaluation protocol reflects a trade-off between scalability and human validation. While this approach enables consistent and reproducible assessment across a large benchmark, it may inherit biases or blind spots from the judge models themselves. Although majority voting across multiple judges mitigates some of these concerns, future work should further investigate alignment between automated judgments and expert human evaluation, particularly for nuanced or ambiguous financial questions.

Overall, our dataset exposes a clear gap between strong single-step extraction performance and fragile multi-step reasoning in financial contexts. Addressing this gap will likely require advances beyond model scaling, including improved training objectives, explicit uncertainty modeling, and mechanisms for error detection and correction in multi-turn interactions.

%% file: Limitations.tex
Our benchmark focuses exclusively on French-language investment documents, which limits the direct generalizability of our findings to other languages or regulatory settings. While this choice addresses a clear gap, financial disclosure practices may differ across jurisdictions.
The benchmark primarily evaluates information extraction and reasoning grounded in explicit document content. It does not cover more speculative or advisory use cases, such as portfolio recommendation or forward-looking decision-making, and should therefore be viewed as assessing foundational document understanding rather than full financial expertise.
Dataset construction relies in part on semi-automatic question generation using large language models, followed by expert revision. We observed occasional references to information outside the provided input context, as well as limited originality and lexical diversity in generated questions, suggesting potential memorization effects and a bias toward easily extractable facts. These observations stress the continued necessity of human expert validation to ensure proper grounding and question quality.
Finally, our evaluation relies on an LLM-as-judge protocol rather than exhaustive human annotation. While majority voting across multiple judges improves robustness, subtle numerical or legal errors may still be missed. In addition, our conversational evaluation highlights indeed error propagation but does not explicitly model uncertainty awareness or error correction.

%% file: appendix_en.tex
\subsection{Dataset Examples}
\label{app:dataset_examples}

This section provides representative examples from the \ScribeFinance benchmark across different task categories. For convenience, all examples have been translated into English, the original French prompts are available in the paper's accompanying repository.

\subsubsection{Table Comprehension Example}
\label{app:table_example}

Figure~\ref{fig:table_example} presents a typical table comprehension task from the dataset.

\begin{figure}[ht!]
\centering
\includegraphics[width=1\columnwidth]{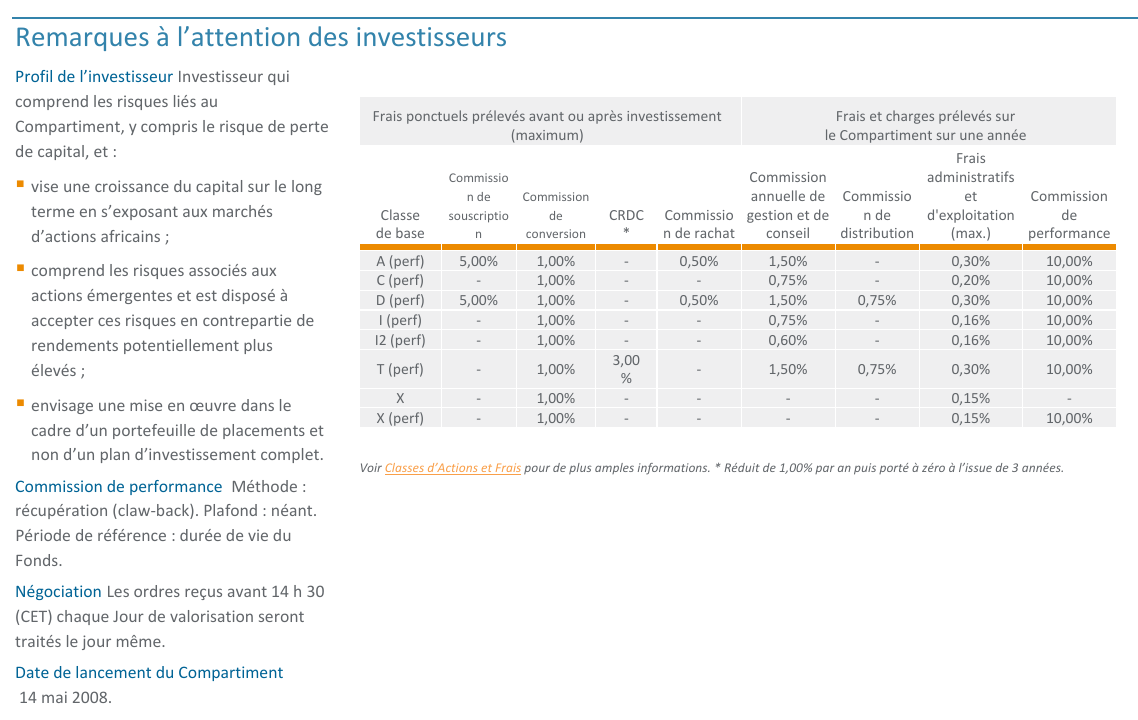}
\caption{Example table from a financial document.}
\label{fig:table_example_image}
\end{figure}

\begin{figure}[ht!]
\centering
\fbox{
\parbox{0.92\columnwidth}{
\textbf{Question:} Which class applies a redemption fee? \\[0.3em]
\textbf{Answer:} A (perf), D (perf)
}
}
\caption{Table comprehension example based on Figure~\ref{fig:table_example_image}.}
\label{fig:table_example}
\end{figure}

\subsubsection{Chart Interpretation Example}
\label{app:chart_example}

Figure~\ref{fig:chart_example} illustrates a chart interpretation task.

\begin{figure}[ht!]
\centering
\includegraphics[width=1\columnwidth]{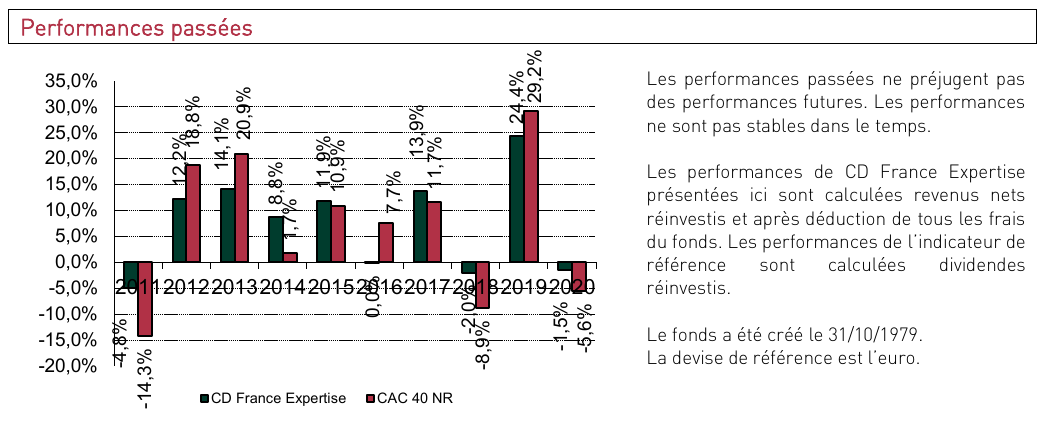}
\caption{Example financial chart requiring visual interpretation.}
\label{fig:chart_example_image}
\end{figure}

\begin{figure}[ht!]
\centering
\fbox{
\parbox{0.92\columnwidth}{
\textbf{Question:} What was the performance of the CD France Expertise fund in 2018? \\[0.3em]
\textbf{Answer:} -2.0\%
}
}
\caption{Chart interpretation example based on Figure~\ref{fig:chart_example_image}.}
\label{fig:chart_example}
\end{figure}

\subsubsection{Conversational Task Example}
\label{app:conv_example}

\begin{figure}[ht!]
\centering
\includegraphics[width=0.98\columnwidth]{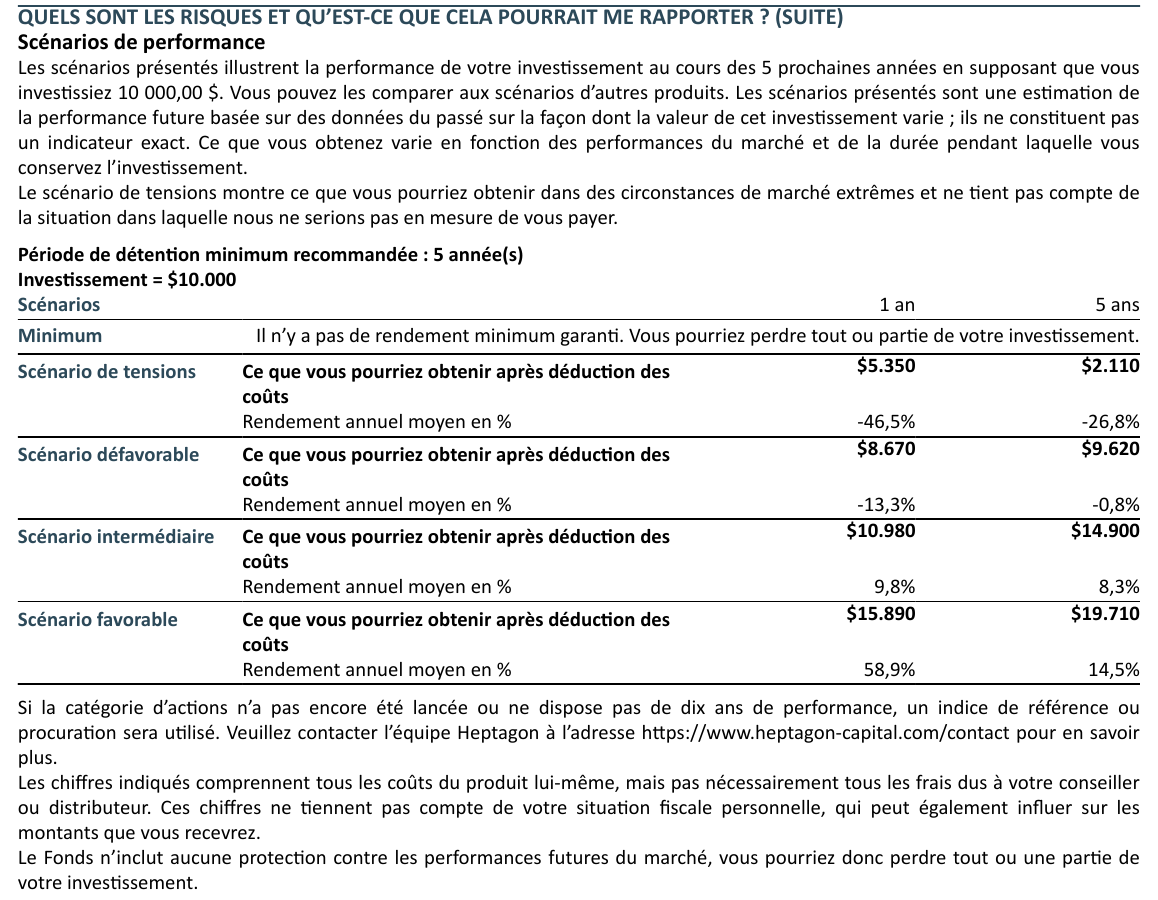}
\caption{Financial projection table showing investment scenarios over different time periods. This image serves as the visual context for the conversational dialogue in Figure~\ref{fig:conv_example}.}
\label{fig:conv_example_image}
\end{figure}

Figure~\ref{fig:conv_example_image} shows the financial projection table used as context for the multi-turn dialogue presented in Figure~\ref{fig:conv_example}. This example demonstrates how questions require maintaining context across turns and performing numerical reasoning based on tabular financial data.

\begin{figure}[ht!]
\centering
\fbox{
\parbox{0.92\columnwidth}{
\textbf{Context:} Financial projection table (Figure~\ref{fig:conv_example_image}) \\[0.5em]
\hrule
\vspace{0.3em}
\textbf{Q1:} If I invest \$25,000 and the favorable scenario occurs after one year, approximately how much will I get at the end of the year? \\[0.2em]
\hspace{1em}A. \$29,500 \\
\hspace{1em}B. \$34,725 \\
\hspace{1em}C. \$39,725 \\
\hspace{1em}D. \$40,000 \\[0.3em]
\textbf{Answer:} C \\[0.5em]
\hrule
\vspace{0.3em}
\textbf{Q2:} And out of curiosity, by how much does that value exceed the intermediate scenario over the same period? \\[0.2em]
\hspace{1em}A. Approximately \$7,625 \\
\hspace{1em}B. A little over \$5,000 \\
\hspace{1em}C. \$12,275 \\
\hspace{1em}D. They are equivalent \\[0.3em]
\textbf{Answer:} C \\[0.5em]
\hrule
\vspace{0.3em}
\textbf{Q3:} OK, but if I look ahead 5 years with this same scenario, what final amount do I reach? \\[0.2em]
\hspace{1em}A. \$32,000 \\
\hspace{1em}B. \$37,250 \\
\hspace{1em}C. \$39,750 \\
\hspace{1em}D. \$43,750 \\[0.3em]
\textbf{Answer:} B \\[0.5em]
\hrule
\vspace{0.3em}
\textbf{Q4:} Oh right, and with the stress scenario over 1 year with the amount I invested\ldots{} roughly how much do I lose? \\[0.2em]
\hspace{1em}A. \$9,250 \\
\hspace{1em}B. \$12,750 \\
\hspace{1em}C. \$11,625 \\
\hspace{1em}D. \$19,650 \\[0.3em]
\textbf{Answer:} C
}
}
\caption{Example of a multi-turn conversational question sequence requiring numerical reasoning across different investment scenarios. Each question builds on previous context, testing the model's ability to maintain coherence and perform calculations based on the tabular financial data shown in Figure~\ref{fig:conv_example_image}.}
\label{fig:conv_example}
\end{figure}

\subsubsection{Translation of Questions from Table \ref{tab:scribe_examples}}
\label{subsubsec:translation_scribe_examples}

Table \ref{tab:transation_scribe_examples} presents the English translations of the example questions shown in Table \ref{tab:scribe_examples}.

\begin{table*}[ht]
    \centering
    {\footnotesize
    \begin{tabular}{cp{12cm}}
\hline
\textbf{Question Type} & \multicolumn{1}{c}{\textbf{Example}} \\
\hline
Text Question & \emph{«~Is this fund actively managed, or does it passively follow an index?~»} \\
Table Comprehension & \emph{«~What are the annual ongoing charges applied by the FCPE?~»} \\
Chart Interpretation &  \emph{«~How many consecutive periods without crystallization are visible in this chart?~»}  \\
Special Cases & \emph{«~Which specific derivative instruments may be used by the sub-fund?~»} \\
    Conversational & 1\textsuperscript{st} turn: \emph{«~If I invest €25,000 in share class A, what would be the maximum entry fees?~»} \\ 
 & 2\textsuperscript{nd} turn:
\emph{«~And if I invest the same amount in share class I or R, would there be any difference in the fees?~»} \\
 \hline
    \end{tabular}
    }
    \caption{English translations of the example questions from the \ScribeFinance dataset, presented in Table \ref{tab:scribe_examples}.}
    \label{tab:transation_scribe_examples}
\end{table*}

\subsection{Prompt Templates}
\label{app:prompt_template}

We used three prompt templates to obtain answers from the LLM during evaluation, depending on the task type. In all cases, models completed an assistant turn prefilled with ``Answer:''. The original prompts were written in French, here we provide their English versions for convenience.

\subsubsection{Image-Based Tasks}
\label{app:prompt_image}

Figure~\ref{fig:prompt_image} presents the template used for Table, Table Yes/No and True/False, Charts, and Special Cases tasks. The model receives an image followed by the question and must complete the assistant turn.

\begin{figure}[ht!]
\centering
\fbox{
\parbox{0.92\columnwidth}{
\textbf{User:} \\[0.3em]
\hspace{1em}\texttt{<image>} \\[0.5em]
\hspace{1em}Question: \{question\} \\
\hspace{1em}Answer the question concisely based on the image provided. Don't include any explanations. \\
\hrule
\vspace{0.3em}
\textbf{Assistant:} \\[0.3em]
\hspace{1em}Answer: \textit{[model completion]}
}
}
\caption{Prompt template for Image-based tasks.}
\label{fig:prompt_image}
\end{figure}

\subsubsection{Text-Based Task}
\label{app:prompt_ner}

Figure~\ref{fig:prompt_ner} presents the template used for the Text Question task, which operates on textual context rather than images.

\begin{figure}[ht!]
\centering
\fbox{
\parbox{0.92\columnwidth}{
\textbf{User:} \\[0.3em]
\hspace{1em}Context: \{context\} \\[0.3em]
\hspace{1em}Question: \{question\} \\
\hspace{1em}Answer the question concisely based on the context provided. Don't include any explanations. \\
\hrule
\vspace{0.3em}
\textbf{Assistant:} \\[0.3em]
\hspace{1em}Answer: \textit{[model completion]}
}
}
\caption{Prompt template for Text-based task.}
\label{fig:prompt_ner}
\end{figure}

\subsubsection{Conversational Tasks}
\label{app:prompt_conv}

Figure~\ref{fig:prompt_conv} illustrates the template used for conversational tasks. The dialogue is built incrementally over 5--10 turns: the image is provided only in the first turn, and each subsequent question is appended as a new user message. In the \textit{Conv.} setting, the model's own previous answers are included in the conversation history. In the \textit{Conv. Gold} setting, ground-truth answers replace model completions.

\begin{figure}[ht!]
\centering
\fbox{
\parbox{0.92\columnwidth}{
\textbf{Turn 1 --- User:} \\[0.3em]
\hspace{1em}\texttt{<image>} \\[0.3em]
\hspace{1em}Answer all questions concisely based on the image provided. Don't include any explanations. \\\{question\_1\}\\[0.5em]
\textbf{Turn 1 --- Assistant:} \\[0.3em]
\hspace{1em}Answer: \textit{[model completion]} \\[0.5em]
\hrule
\vspace{0.3em}
\textbf{Turn 2 --- User:} \\[0.3em]
\hspace{1em}\{question\_2\} \\[0.3em]
\textbf{Turn 2 --- Assistant:} \\[0.3em]
\hspace{1em}Answer: \textit{[model completion]} \\[0.5em]
\hrule
\vspace{0.3em}
\hspace{1em}\textit{... continued for all turns ...}
}
}
\caption{Prompt template for conversational tasks. The image is provided once at the first turn; subsequent turns contain only the question. In the \textit{Conv. Gold} setting, ground-truth answers replace model completions in the history.}
\label{fig:prompt_conv}
\end{figure}

\subsubsection{LLM-as-judge Evaluation Template}
\label{app:prompt_eval}

Figure~\ref{fig:prompt_eval} presents the prompt template used for LLM-as-judge evaluation. Each judge receives the question, reference answer, and prediction, then outputs ``Correct'' or ``Incorrect''.

\begin{figure}[ht]
\centering
\small
\fbox{
\parbox{0.94\columnwidth}{
\textbf{User:} \\[0.2em]
You will receive a question, a reference answer, and an answer to evaluate. Your task is to determine whether the prediction is correct or incorrect. \\[0.3em]
Evaluation rules: \\
1. A prediction is correct if it accurately answers the question based on the reference answer. \\
2. If the answer involves a numerical or financial value, consider as equivalent any expressions representing the same value (e.g., 20\% = 0.2; 1,000,000 = 1 million; 2,200,000 = 2.2M; 12.3 = 12.3). \\
3. If the question is multiple-choice, an answer is correct if it matches exactly one of the correct options (by letter, number, or text). \\
4. If the prediction is an exact paraphrase of the reference, it is correct. \\
5. Do not take into account phrasing or style, only factual or numerical accuracy. \\
6. Respond only with "Correct" or "Incorrect". \\[0.3em]
Examples: \\
Question: \{example question\}\\
Reference: \{example reference\}\\
Answer to evaluate: \{example prediction\}
Expected evaluation: \{example expected evaluation\}\\
\textit{[5 examples total of numerical and MCQ cases with correct and incorrect answer]} \\[0.3em]
Answer to evaluate: \\
Question: \{question\} \\
Reference answer: \{reference\} \\
Answer to evaluate: \{prediction\} \\[0.4em]
\hrule
\vspace{0.2em}
\textbf{Assistant:} \\
Evaluation: \textit{[model completion]}
}
}
\caption{LLM-as-judge evaluation template. Five examples covering numerical equivalence and multiple-choice formats are included in the full prompt.}
\label{fig:prompt_eval}
\end{figure}

%% file: Scribe_Models_evaluation_2.bib
@inproceedings{lithgow-serrano-etal-2025-assessing,
    title = "Assessing {RAG} System Capabilities on Financial Documents",
    author = "Lithgow-Serrano, Oscar  and
      Kletz, David  and
      Kanjirangat, Vani  and
      Adametz, David  and
      Lunghi, Marzio  and
      Bonesana, Claudio  and
      Tristany-Farinha, Matilde  and
      Li, Yuntao  and
      Repplinger, Detlef  and
      Pierbattista, Marco  and
      Stan, Stefania  and
      Szehr, Oleg",
    editor = "Chen, Chung-Chi  and
      Winata, Genta Indra  and
      Rawls, Stephen  and
      Das, Anirban  and
      Chen, Hsin-Hsi  and
      Takamura, Hiroya",
    booktitle = "Proceedings of The 10th Workshop on Financial Technology and Natural Language Processing",
    month = nov,
    year = "2025",
    address = "Suzhou, China",
    publisher = "Association for Computational Linguistics",
    url = "https://aclanthology.org/2025.finnlp-2.9/",
    doi = "10.18653/v1/2025.finnlp-2.9",
    pages = "124--147"
}

@inproceedings{chen-etal-2025-beyond,
    title = "Beyond the Surface: Measuring Self-Preference in {LLM} Judgments",
    author = "Chen, Zhi-Yuan  and
      Wang, Hao  and
      Zhang, Xinyu  and
      Hu, Enrui  and
      Lin, Yankai",
    editor = "Christodoulopoulos, Christos  and
      Chakraborty, Tanmoy  and
      Rose, Carolyn  and
      Peng, Violet",
    booktitle = "Proceedings of the 2025 Conference on Empirical Methods in Natural Language Processing",
    month = nov,
    year = "2025",
    address = "Suzhou, China",
    publisher = "Association for Computational Linguistics",
    url = "https://aclanthology.org/2025.emnlp-main.86/",
    doi = "10.18653/v1/2025.emnlp-main.86",
    pages = "1653--1672",
    ISBN = "979-8-89176-332-6"
}

@inproceedings{Zheng2023llmasajudge,
 author = {Zheng, Lianmin and Chiang, Wei-Lin and Sheng, Ying and Zhuang, Siyuan and Wu, Zhanghao and Zhuang, Yonghao and Lin, Zi and Li, Zhuohan and Li, Dacheng and Xing, Eric and Zhang, Hao and Gonzalez, Joseph E and Stoica, Ion},
 booktitle = {Advances in Neural Information Processing Systems},
 editor = {A. Oh and T. Naumann and A. Globerson and K. Saenko and M. Hardt and S. Levine},
 pages = {46595--46623},
 publisher = {Curran Associates, Inc.},
 title = {Judging LLM-as-a-Judge with MT-Bench and Chatbot Arena},
 url = {https://proceedings.neurips.cc/paper_files/paper/2023/file/91f18a1287b398d378ef22505bf41832-Paper-Datasets_and_Benchmarks.pdf},
 volume = {36},
 year = {2023}
}

@misc{faysse2025colpaliefficientdocumentretrieval,
      title={ColPali: Efficient Document Retrieval with Vision Language Models}, 
      author={Manuel Faysse and Hugues Sibille and Tony Wu and Bilel Omrani and Gautier Viaud and Céline Hudelot and Pierre Colombo},
      year={2025},
      eprint={2407.01449},
      archivePrefix={arXiv},
      primaryClass={cs.IR},
      url={https://arxiv.org/abs/2407.01449}, 
}

@misc{xue2025fammabenchmarkfinancialdomain,
      title={FAMMA: A Benchmark for Financial Domain Multilingual Multimodal Question Answering}, 
      author={Siqiao Xue and Xiaojing Li and Fan Zhou and Qingyang Dai and Zhixuan Chu and Hongyuan Mei},
      year={2025},
      eprint={2410.04526},
      archivePrefix={arXiv},
      primaryClass={cs.CL},
      url={https://arxiv.org/abs/2410.04526}, 
}

@misc{g20_2009_london_statement,
  author       = {{G20}},
  title        = {Leaders' Statement: The Global Plan for Recovery and Reform},
  year         = {2009},
  howpublished = {G20 London Summit, April 2009},
  url          = {https://www.g20.utoronto.ca/analysis/hse/2009-london-compliance.pdf}
}

@misc{ec_2014_mifid2,
  author       = {{European Commission}},
  title        = {Directive 2014/65/EU of the European Parliament and of the Council of 15 May 2014 on Markets in Financial Instruments (MiFID II)},
  year         = {2014},
  howpublished = {Official Journal of the European Union},
  url          = {https://eur-lex.europa.eu/legal-content/EN/TXT/?uri=CELEX%3A02014L0065-20250117}
}

@misc{eu_2014_priips,
  author       = {{European Union}},
  title        = {Regulation (EU) No 1286/2014 of the European Parliament and of the Council of 26 November 2014 on Key Information Documents for Packaged Retail and Insurance-Based Investment Products (PRIIPs)},
  year         = {2014},
  howpublished = {Official Journal of the European Union},
  url          = {https://eur-lex.europa.eu/legal-content/EN/TXT/?uri=CELEX%3A02014R1286-20240109}
}

@inproceedings{dhoffschmidt-etal-2020-fquad,
    title = "{FQ}u{AD}: {F}rench Question Answering Dataset",
    author = "d{'}Hoffschmidt, Martin  and
      Belblidia, Wacim  and
      Heinrich, Quentin  and
      Brendl{\'e}, Tom  and
      Vidal, Maxime",
    editor = "Cohn, Trevor  and
      He, Yulan  and
      Liu, Yang",
    booktitle = "Findings of the Association for Computational Linguistics: EMNLP 2020",
    month = nov,
    year = "2020",
    address = "Online",
    publisher = "Association for Computational Linguistics",
    url = "https://aclanthology.org/2020.findings-emnlp.107/",
    doi = "10.18653/v1/2020.findings-emnlp.107",
    pages = "1193--1208"
}

@misc{labrak_frenchmedmcqa_2023,
	title = {{FrenchMedMCQA}: {A} {French} {Multiple}-{Choice} {Question} {Answering} {Dataset} for {Medical} domain},
	shorttitle = {{FrenchMedMCQA}},
	url = {http://arxiv.org/abs/2304.04280},
	doi = {10.48550/arXiv.2304.04280},
	urldate = {2025-05-15},
	publisher = {arXiv},
	author = {Labrak, Yanis and Bazoge, Adrien and Dufour, Richard and Rouvier, Mickael and Morin, Emmanuel and Daille, Béatrice and Gourraud, Pierre-Antoine},
	month = apr,
	year = {2023},
	note = {arXiv:2304.04280 [cs]},
}

@misc{nangia_crows-pairs_2020,
	title = {{CrowS}-{Pairs}: {A} {Challenge} {Dataset} for {Measuring} {Social} {Biases} in {Masked} {Language} {Models}},
	shorttitle = {{CrowS}-{Pairs}},
	url = {http://arxiv.org/abs/2010.00133},
	doi = {10.48550/arXiv.2010.00133},
	urldate = {2025-05-15},
	publisher = {arXiv},
	author = {Nangia, Nikita and Vania, Clara and Bhalerao, Rasika and Bowman, Samuel R.},
	month = sep,
	year = {2020},
	note = {arXiv:2010.00133 [cs]},
}

@misc{keraron_project_2020,
	title = {Project {PIAF}: {Building} a {Native} {French} {Question}-{Answering} {Dataset}},
	shorttitle = {Project {PIAF}},
	url = {http://arxiv.org/abs/2007.00968},
	doi = {10.48550/arXiv.2007.00968},
	urldate = {2025-05-15},
	publisher = {arXiv},
	author = {Keraron, Rachel and Lancrenon, Guillaume and Bras, Mathilde and Allary, Frédéric and Moyse, Gilles and Scialom, Thomas and Soriano-Morales, Edmundo-Pavel and Staiano, Jacopo},
	month = jul,
	year = {2020},
	note = {arXiv:2007.00968 [cs]},
}

@misc{longpre_mkqa_2021,
	title = {{MKQA}: {A} {Linguistically} {Diverse} {Benchmark} for {Multilingual} {Open} {Domain} {Question} {Answering}},
	shorttitle = {{MKQA}},
	url = {http://arxiv.org/abs/2007.15207},
	doi = {10.48550/arXiv.2007.15207},
	urldate = {2025-05-15},
	publisher = {arXiv},
	author = {Longpre, Shayne and Lu, Yi and Daiber, Joachim},
	month = aug,
	year = {2021},
	note = {arXiv:2007.15207 [cs]},
}

@misc{zhu_tat-qa_2021,
	title = {{TAT}-{QA}: {A} {Question} {Answering} {Benchmark} on a {Hybrid} of {Tabular} and {Textual} {Content} in {Finance}},
	shorttitle = {{TAT}-{QA}},
	url = {http://arxiv.org/abs/2105.07624},
	doi = {10.48550/arXiv.2105.07624},
	urldate = {2025-05-15},
	publisher = {arXiv},
	author = {Zhu, Fengbin and Lei, Wenqiang and Huang, Youcheng and Wang, Chao and Zhang, Shuo and Lv, Jiancheng and Feng, Fuli and Chua, Tat-Seng},
	month = jun,
	year = {2021},
	note = {arXiv:2105.07624 [cs]},
}

@misc{chen_finqa_2022,
	title = {{FinQA}: {A} {Dataset} of {Numerical} {Reasoning} over {Financial} {Data}},
	shorttitle = {{FinQA}},
	url = {http://arxiv.org/abs/2109.00122},
	doi = {10.48550/arXiv.2109.00122},
	urldate = {2025-05-15},
	publisher = {arXiv},
	author = {Chen, Zhiyu and Chen, Wenhu and Smiley, Charese and Shah, Sameena and Borova, Iana and Langdon, Dylan and Moussa, Reema and Beane, Matt and Huang, Ting-Hao and Routledge, Bryan and Wang, William Yang},
	month = may,
	year = {2022},
	note = {arXiv:2109.00122 [cs]},
}

@misc{chen_convfinqa_2022,
	title = {{ConvFinQA}: {Exploring} the {Chain} of {Numerical} {Reasoning} in {Conversational} {Finance} {Question} {Answering}},
	shorttitle = {{ConvFinQA}},
	url = {http://arxiv.org/abs/2210.03849},
	doi = {10.48550/arXiv.2210.03849},
	urldate = {2025-05-15},
	publisher = {arXiv},
	author = {Chen, Zhiyu and Li, Shiyang and Smiley, Charese and Ma, Zhiqiang and Shah, Sameena and Wang, William Yang},
	month = oct,
	year = {2022},
	note = {arXiv:2210.03849 [cs]},
}

@misc{deng_pacific_2023,
	title = {{PACIFIC}: {Towards} {Proactive} {Conversational} {Question} {Answering} over {Tabular} and {Textual} {Data} in {Finance}},
	shorttitle = {{PACIFIC}},
	url = {http://arxiv.org/abs/2210.08817},
	doi = {10.48550/arXiv.2210.08817},
	urldate = {2025-05-15},
	publisher = {arXiv},
	author = {Deng, Yang and Lei, Wenqiang and Zhang, Wenxuan and Lam, Wai and Chua, Tat-Seng},
	month = mar,
	year = {2023},
	note = {arXiv:2210.08817 [cs]},
}

@inproceedings{liu_xqa_2019,
	address = {Florence, Italy},
	title = {{XQA}: {A} {Cross}-lingual {Open}-domain {Question} {Answering} {Dataset}},
	shorttitle = {{XQA}},
	url = {https://www.aclweb.org/anthology/P19-1227},
	doi = {10.18653/v1/P19-1227},
	language = {en},
	urldate = {2025-05-15},
	booktitle = {Proceedings of the 57th {Annual} {Meeting} of the {Association} for {Computational} {Linguistics}},
	publisher = {Association for Computational Linguistics},
	author = {Liu, Jiahua and Lin, Yankai and Liu, Zhiyuan and Sun, Maosong},
	year = {2019},
	pages = {2358--2368},
}

@article{zhang_miracl_2023,
	title = {\textbf{{MIRACL}} : {A} {Multilingual} {Retrieval} {Dataset} {Covering} 18 {Diverse} {Languages}},
	volume = {11},
	issn = {2307-387X},
	shorttitle = {\textbf{{MIRACL}}},
	url = {https://direct.mit.edu/tacl/article/doi/10.1162/tacl_a_00595/117438/MIRACL-A-Multilingual-Retrieval-Dataset-Covering},
	doi = {10.1162/tacl_a_00595},
	language = {en},
	urldate = {2025-05-15},
	journal = {Transactions of the Association for Computational Linguistics},
	author = {Zhang, Xinyu and Thakur, Nandan and Ogundepo, Odunayo and Kamalloo, Ehsan and Alfonso-Hermelo, David and Li, Xiaoguang and Liu, Qun and Rezagholizadeh, Mehdi and Lin, Jimmy},
	month = sep,
	year = {2023},
	pages = {1114--1131},
}

@article{neveol_french_2022,
	title = {French {CrowS}-{Pairs}: {Extension} à une langue autre que l’anglais d’un corpus de mesure des biais sociétaux dans les modèles de langue masqués},
	journal = {Actes de la 29e Conférence sur le Traitement Automatique des Langues Naturelles},
	author = {Névéol, Aurélie and Dupont, Yoann and Bezançon, Julien and Fort, Karën},
	year = {2022},
}

@misc{qwen3technicalreport,
      title={Qwen3 Technical Report}, 
      author={{Qwen Team}},
      year={2025},
      eprint={2505.09388},
      archivePrefix={arXiv},
      primaryClass={cs.CL},
      url={https://arxiv.org/abs/2505.09388}, 
}

@misc{gemma_2025,
      title={Gemma 3 Technical Report}, 
      author={{Gemma Team} and Aishwarya Kamath and Johan Ferret and Shreya Pathak and Nino Vieillard and others},
      year={2025},
      eprint={2503.19786},
      archivePrefix={arXiv},
      primaryClass={cs.CL},
      url={https://arxiv.org/abs/2503.19786}, 
}

@inproceedings{rajpurkar2016squad,
  title={SQuAD: 100,000+ Questions for Machine Comprehension of Text},
  author={Rajpurkar, Pranav and Zhang, Jian and Lopyrev, Konstantin and Liang, Percy},
  booktitle={Proceedings of the 2016 Conference on Empirical Methods in Natural Language Processing},
  year={2016},
  organization={Association for Computational Linguistics}
}

@article{clark2026molmo2,
  title={Molmo2: Open Weights and Data for Vision-Language Models with Video Understanding and Grounding},
  author={Clark, Christopher and Zhang, Jieyu and Ma, Zixian and Park, Jae Sung and Salehi, Mohammadreza and Tripathi, Rohun and Lee, Sangho and Ren, Zhongzheng and Kim, Chris Dongjoo and Yang, Yinuo and others},
  journal={arXiv preprint arXiv:2601.10611},
  year={2026}
}

@misc{grattafiori2024llama3herdmodels,
      title={The Llama 3 Herd of Models}, 
      author={Aaron Grattafiori and Abhimanyu Dubey and Abhinav Jauhri and Abhinav Pandey and Abhishek Kadian and Ahmad Al-Dahle and Aiesha Letman and Akhil Mathur and Alan Schelten and Alex Vaughan and Amy Yang and Angela Fan and Anirudh Goyal and Anthony Hartshorn and Aobo Yang and Archi Mitra and Archie Sravankumar and Artem Korenev and Arthur Hinsvark and Arun Rao and Aston Zhang and Aurelien Rodriguez and Austen Gregerson and Ava Spataru and Baptiste Roziere and Bethany Biron and Binh Tang and Bobbie Chern and Charlotte Caucheteux and Chaya Nayak and Chloe Bi and Chris Marra and Chris McConnell and Christian Keller and Christophe Touret and Chunyang Wu and Corinne Wong and Cristian Canton Ferrer and Cyrus Nikolaidis and Damien Allonsius and Daniel Song and Danielle Pintz and Danny Livshits and Danny Wyatt and David Esiobu and Dhruv Choudhary and Dhruv Mahajan and Diego Garcia-Olano and Diego Perino and Dieuwke Hupkes and Egor Lakomkin and Ehab AlBadawy and Elina Lobanova and Emily Dinan and Eric Michael Smith and Filip Radenovic and Francisco Guzmán and Frank Zhang and Gabriel Synnaeve and Gabrielle Lee and Georgia Lewis Anderson and Govind Thattai and Graeme Nail and Gregoire Mialon and Guan Pang and Guillem Cucurell and Hailey Nguyen and Hannah Korevaar and Hu Xu and Hugo Touvron and Iliyan Zarov and Imanol Arrieta Ibarra and Isabel Kloumann and Ishan Misra and Ivan Evtimov and Jack Zhang and Jade Copet and Jaewon Lee and Jan Geffert and Jana Vranes and Jason Park and Jay Mahadeokar and Jeet Shah and Jelmer van der Linde and Jennifer Billock and Jenny Hong and Jenya Lee and Jeremy Fu and Jianfeng Chi and Jianyu Huang and Jiawen Liu and Jie Wang and Jiecao Yu and Joanna Bitton and Joe Spisak and Jongsoo Park and Joseph Rocca and Joshua Johnstun and Joshua Saxe and Junteng Jia and Kalyan Vasuden Alwala and Karthik Prasad and Kartikeya Upasani and Kate Plawiak and Ke Li and Kenneth Heafield and Kevin Stone and Khalid El-Arini and Krithika Iyer and Kshitiz Malik and Kuenley Chiu and Kunal Bhalla and Kushal Lakhotia and Lauren Rantala-Yeary and Laurens van der Maaten and Lawrence Chen and Liang Tan and Liz Jenkins and Louis Martin and Lovish Madaan and Lubo Malo and Lukas Blecher and Lukas Landzaat and Luke de Oliveira and Madeline Muzzi and Mahesh Pasupuleti and Mannat Singh and Manohar Paluri and Marcin Kardas and Maria Tsimpoukelli and Mathew Oldham and Mathieu Rita and Maya Pavlova and Melanie Kambadur and Mike Lewis and Min Si and Mitesh Kumar Singh and Mona Hassan and Naman Goyal and Narjes Torabi and Nikolay Bashlykov and Nikolay Bogoychev and Niladri Chatterji and Ning Zhang and Olivier Duchenne and Onur Çelebi and Patrick Alrassy and Pengchuan Zhang and Pengwei Li and Petar Vasic and Peter Weng and Prajjwal Bhargava and Pratik Dubal and Praveen Krishnan and Punit Singh Koura and Puxin Xu and Qing He and Qingxiao Dong and Ragavan Srinivasan and Raj Ganapathy and Ramon Calderer and Ricardo Silveira Cabral and Robert Stojnic and Roberta Raileanu and Rohan Maheswari and Rohit Girdhar and Rohit Patel and Romain Sauvestre and Ronnie Polidoro and Roshan Sumbaly and Ross Taylor and Ruan Silva and Rui Hou and Rui Wang and Saghar Hosseini and Sahana Chennabasappa and Sanjay Singh and Sean Bell and Seohyun Sonia Kim and Sergey Edunov and Shaoliang Nie and Sharan Narang and Sharath Raparthy and Sheng Shen and Shengye Wan and Shruti Bhosale and Shun Zhang and Simon Vandenhende and Soumya Batra and Spencer Whitman and Sten Sootla and Stephane Collot and Suchin Gururangan and Sydney Borodinsky and Tamar Herman and Tara Fowler and Tarek Sheasha and Thomas Georgiou and Thomas Scialom and Tobias Speckbacher and Todor Mihaylov and Tong Xiao and Ujjwal Karn and Vedanuj Goswami and Vibhor Gupta and Vignesh Ramanathan and Viktor Kerkez and Vincent Gonguet and Virginie Do and Vish Vogeti and Vítor Albiero and Vladan Petrovic and Weiwei Chu and Wenhan Xiong and Wenyin Fu and Whitney Meers and Xavier Martinet and Xiaodong Wang and Xiaofang Wang and Xiaoqing Ellen Tan and Xide Xia and Xinfeng Xie and Xuchao Jia and Xuewei Wang and Yaelle Goldschlag and Yashesh Gaur and Yasmine Babaei and Yi Wen and Yiwen Song and Yuchen Zhang and Yue Li and Yuning Mao and Zacharie Delpierre Coudert and Zheng Yan and Zhengxing Chen and Zoe Papakipos and Aaditya Singh and Aayushi Srivastava and Abha Jain and Adam Kelsey and Adam Shajnfeld and Adithya Gangidi and Adolfo Victoria and Ahuva Goldstand and Ajay Menon and Ajay Sharma and Alex Boesenberg and Alexei Baevski and Allie Feinstein and Amanda Kallet and Amit Sangani and Amos Teo and Anam Yunus and Andrei Lupu and Andres Alvarado and Andrew Caples and Andrew Gu and Andrew Ho and Andrew Poulton and Andrew Ryan and Ankit Ramchandani and Annie Dong and Annie Franco and Anuj Goyal and Aparajita Saraf and Arkabandhu Chowdhury and Ashley Gabriel and Ashwin Bharambe and Assaf Eisenman and Azadeh Yazdan and Beau James and Ben Maurer and Benjamin Leonhardi and Bernie Huang and Beth Loyd and Beto De Paola and Bhargavi Paranjape and Bing Liu and Bo Wu and Boyu Ni and Braden Hancock and Bram Wasti and Brandon Spence and Brani Stojkovic and Brian Gamido and Britt Montalvo and Carl Parker and Carly Burton and Catalina Mejia and Ce Liu and Changhan Wang and Changkyu Kim and Chao Zhou and Chester Hu and Ching-Hsiang Chu and Chris Cai and Chris Tindal and Christoph Feichtenhofer and Cynthia Gao and Damon Civin and Dana Beaty and Daniel Kreymer and Daniel Li and David Adkins and David Xu and Davide Testuggine and Delia David and Devi Parikh and Diana Liskovich and Didem Foss and Dingkang Wang and Duc Le and Dustin Holland and Edward Dowling and Eissa Jamil and Elaine Montgomery and Eleonora Presani and Emily Hahn and Emily Wood and Eric-Tuan Le and Erik Brinkman and Esteban Arcaute and Evan Dunbar and Evan Smothers and Fei Sun and Felix Kreuk and Feng Tian and Filippos Kokkinos and Firat Ozgenel and Francesco Caggioni and Frank Kanayet and Frank Seide and Gabriela Medina Florez and Gabriella Schwarz and Gada Badeer and Georgia Swee and Gil Halpern and Grant Herman and Grigory Sizov and Guangyi and Zhang and Guna Lakshminarayanan and Hakan Inan and Hamid Shojanazeri and Han Zou and Hannah Wang and Hanwen Zha and Haroun Habeeb and Harrison Rudolph and Helen Suk and Henry Aspegren and Hunter Goldman and Hongyuan Zhan and Ibrahim Damlaj and Igor Molybog and Igor Tufanov and Ilias Leontiadis and Irina-Elena Veliche and Itai Gat and Jake Weissman and James Geboski and James Kohli and Janice Lam and Japhet Asher and Jean-Baptiste Gaya and Jeff Marcus and Jeff Tang and Jennifer Chan and Jenny Zhen and Jeremy Reizenstein and Jeremy Teboul and Jessica Zhong and Jian Jin and Jingyi Yang and Joe Cummings and Jon Carvill and Jon Shepard and Jonathan McPhie and Jonathan Torres and Josh Ginsburg and Junjie Wang and Kai Wu and Kam Hou U and Karan Saxena and Kartikay Khandelwal and Katayoun Zand and Kathy Matosich and Kaushik Veeraraghavan and Kelly Michelena and Keqian Li and Kiran Jagadeesh and Kun Huang and Kunal Chawla and Kyle Huang and Lailin Chen and Lakshya Garg and Lavender A and Leandro Silva and Lee Bell and Lei Zhang and Liangpeng Guo and Licheng Yu and Liron Moshkovich and Luca Wehrstedt and Madian Khabsa and Manav Avalani and Manish Bhatt and Martynas Mankus and Matan Hasson and Matthew Lennie and Matthias Reso and Maxim Groshev and Maxim Naumov and Maya Lathi and Meghan Keneally and Miao Liu and Michael L. Seltzer and Michal Valko and Michelle Restrepo and Mihir Patel and Mik Vyatskov and Mikayel Samvelyan and Mike Clark and Mike Macey and Mike Wang and Miquel Jubert Hermoso and Mo Metanat and Mohammad Rastegari and Munish Bansal and Nandhini Santhanam and Natascha Parks and Natasha White and Navyata Bawa and Nayan Singhal and Nick Egebo and Nicolas Usunier and Nikhil Mehta and Nikolay Pavlovich Laptev and Ning Dong and Norman Cheng and Oleg Chernoguz and Olivia Hart and Omkar Salpekar and Ozlem Kalinli and Parkin Kent and Parth Parekh and Paul Saab and Pavan Balaji and Pedro Rittner and Philip Bontrager and Pierre Roux and Piotr Dollar and Polina Zvyagina and Prashant Ratanchandani and Pritish Yuvraj and Qian Liang and Rachad Alao and Rachel Rodriguez and Rafi Ayub and Raghotham Murthy and Raghu Nayani and Rahul Mitra and Rangaprabhu Parthasarathy and Raymond Li and Rebekkah Hogan and Robin Battey and Rocky Wang and Russ Howes and Ruty Rinott and Sachin Mehta and Sachin Siby and Sai Jayesh Bondu and Samyak Datta and Sara Chugh and Sara Hunt and Sargun Dhillon and Sasha Sidorov and Satadru Pan and Saurabh Mahajan and Saurabh Verma and Seiji Yamamoto and Sharadh Ramaswamy and Shaun Lindsay and Shaun Lindsay and Sheng Feng and Shenghao Lin and Shengxin Cindy Zha and Shishir Patil and Shiva Shankar and Shuqiang Zhang and Shuqiang Zhang and Sinong Wang and Sneha Agarwal and Soji Sajuyigbe and Soumith Chintala and Stephanie Max and Stephen Chen and Steve Kehoe and Steve Satterfield and Sudarshan Govindaprasad and Sumit Gupta and Summer Deng and Sungmin Cho and Sunny Virk and Suraj Subramanian and Sy Choudhury and Sydney Goldman and Tal Remez and Tamar Glaser and Tamara Best and Thilo Koehler and Thomas Robinson and Tianhe Li and Tianjun Zhang and Tim Matthews and Timothy Chou and Tzook Shaked and Varun Vontimitta and Victoria Ajayi and Victoria Montanez and Vijai Mohan and Vinay Satish Kumar and Vishal Mangla and Vlad Ionescu and Vlad Poenaru and Vlad Tiberiu Mihailescu and Vladimir Ivanov and Wei Li and Wenchen Wang and Wenwen Jiang and Wes Bouaziz and Will Constable and Xiaocheng Tang and Xiaojian Wu and Xiaolan Wang and Xilun Wu and Xinbo Gao and Yaniv Kleinman and Yanjun Chen and Ye Hu and Ye Jia and Ye Qi and Yenda Li and Yilin Zhang and Ying Zhang and Yossi Adi and Youngjin Nam and Yu and Wang and Yu Zhao and Yuchen Hao and Yundi Qian and Yunlu Li and Yuzi He and Zach Rait and Zachary DeVito and Zef Rosnbrick and Zhaoduo Wen and Zhenyu Yang and Zhiwei Zhao and Zhiyu Ma},
      year={2024},
      eprint={2407.21783},
      archivePrefix={arXiv},
      primaryClass={cs.AI},
      url={https://arxiv.org/abs/2407.21783}, 
}

@misc{agrawal2024pixtral12b,
      title={Pixtral 12B}, 
      author={Pravesh Agrawal and Szymon Antoniak and Emma Bou Hanna and Baptiste Bout and Devendra Chaplot and Jessica Chudnovsky and Diogo Costa and Baudouin De Monicault and Saurabh Garg and Theophile Gervet and Soham Ghosh and Amélie Héliou and Paul Jacob and Albert Q. Jiang and Kartik Khandelwal and Timothée Lacroix and Guillaume Lample and Diego Las Casas and Thibaut Lavril and Teven Le Scao and Andy Lo and William Marshall and Louis Martin and Arthur Mensch and Pavankumar Muddireddy and Valera Nemychnikova and Marie Pellat and Patrick Von Platen and Nikhil Raghuraman and Baptiste Rozière and Alexandre Sablayrolles and Lucile Saulnier and Romain Sauvestre and Wendy Shang and Roman Soletskyi and Lawrence Stewart and Pierre Stock and Joachim Studnia and Sandeep Subramanian and Sagar Vaze and Thomas Wang and Sophia Yang},
      year={2024},
      eprint={2410.07073},
      archivePrefix={arXiv},
      primaryClass={cs.CV},
      url={https://arxiv.org/abs/2410.07073}, 
}
